\documentclass[journal]{IEEEtran}
\usepackage{amsmath, amsfonts, amssymb, bm, amsthm}
\usepackage[utf8]{inputenc}
\usepackage{enumitem} 

\usepackage[ruled,vlined]{algorithm2e} 
\makeatletter
\renewcommand\@makefnmark{\hbox{\@textsuperscript{\normalfont\color{black}\@thefnmark}}}
\makeatother

\usepackage[bottom]{footmisc}
\usepackage{color, xcolor, soul, framed}
\usepackage{cancel}
\usepackage[numbers,sort&compress]{natbib}
\usepackage[font=normalfont]{caption} 

\usepackage{subcaption}
\usepackage{booktabs}
\usepackage{mathtools}
\usepackage{graphicx}
\usepackage{hyperref}
\usepackage{varwidth}
\usepackage{amsmath}

\usepackage{multirow}
\usepackage{array}
\usepackage{lipsum}

\definecolor{turquoise}{rgb}{.0,.3,1.0}

\newcommand{\TT}{\operatorname{T}}

\newcommand{\EV}{\operatorname{E}}
\newcommand{\VAR}{\operatorname{Var}}

\newcommand{\DIAG}{\operatorname{diag}}

\hypersetup{
    colorlinks=true,    citecolor=blue,    linkcolor=red,    filecolor=magenta,  urlcolor=blue,    pdftitle={Overleaf Example},
    }

\begin{document}

\title{Towards Learning-Based Gyrocompassing}

\author{Daniel~Engelsman and Itzik~Klein
\thanks{The authors are with the Hatter Department of Marine Technologies, Charney School of Marine Sciences, University of Haifa, Israel.\\ E-mails: \{dengelsm@campus, kitzik@univ\}.haifa.ac.il}}


\maketitle

\begin{abstract}
Inertial navigation systems (INS) are widely used in both manned and autonomous platforms. One of the most critical tasks prior to their operation is to accurately determine their initial alignment while stationary, as it forms the cornerstone for the entire INS operational trajectory. 
While low-performance accelerometers can easily determine roll and pitch angles (leveling), establishing the heading angle (gyrocompassing) with low-performance gyros proves to be a challenging task without additional sensors. This arises from the limited signal strength of Earth's rotation rate, often overridden by gyro noise itself.
To circumvent this deficiency, in this study we present a practical deep learning framework to effectively compensate for the inherent errors in low-performance gyroscopes. The resulting capability enables gyrocompassing, thereby eliminating the need for subsequent prolonged filtering phase (fine alignment).
Through the development of theory and experimental validation, we demonstrate that the improved initial conditions establish a new lower error bound, bringing affordable gyros one step closer to being utilized in high-end tactical tasks.
\end{abstract}


\section{Introduction}
\IEEEPARstart{S}{ince} the dawn of mankind, navigation was made possible through the use of celestial bodies as reliable references points. In the mid-19th century, the Foucault pendulum was introduced, showcasing Earth's rotation through the graceful precession of its swinging plane—a pivotal discovery that enhanced comprehension of the gyroscopic effect within the realm of dynamics. By the end of that century, the adoption of gimbals—mechanical structures designed to isolate carrier motion from the gyroscopes—had paved the way for the utilization of rapidly spinning wheel gyroscopes to maintain a steady reference direction \cite{6782505}.
\\
During the mid-20th century, miniaturization of solid-state electronics gave rise to the creation of precision gyroscopes like the ring laser gyroscope (RLG), dynamically tuned gyroscope (DTG), and fiber optic gyroscope (FOG).
Yet, it was not until the 1990s that chip manufacturers embarked on the production of the celebrated micro electronic mechanical systems (MEMS) technology. The fabricated sensors have gradually transitioned from being auxiliary subcomponents in gaming applications to becoming an integral part of autonomous vehicles, reshaping the landscape of our everyday electronics \cite{ayazi2011multi}.
\\
However despite technological advancements, the pursuit to derive comprehensive angle information from raw inertial measurements has endured, due to its pivotal role in numerous tasks, including navigation, guidance, control, tracking, and path planning.
\\
Acquiring the attitude angles, i.e. the roll and pitch angles, is regarded as a straightforward procedure achieved using a stationary accelerometer. Inexpensive MEMS accelerometer with a 10 milli-g noise density can ascertain sub-degree accuracy thanks to the pronounced gravity signal. 
Nevertheless, determining the initial heading angle—the relative angle between the longitudinal body axis and the North direction—is notably more challenging due to a considerably weaker signal of the Earth's rotation rate. 
\\
One viable solution to tackle the low signal-to-noise ratio (SNR) is magnetic-field-based methods, wherein measurements of Earth's magnetic field lines are employed to infer the heading angle in relation to magnetic north. However, despite their effectiveness and ease of use, their susceptibility to interference from ferromagnetic metals prioritizes them lower when dependable performance is needed. 
Moreover, a calibration for the declination angle is necessary to determine the relative angle between magnetic North and true North. Thereby, the direct acquisition of the latter provides a geographic-independent solution that is robust to electromagnetic effects, relying on a fixed reference frame in the form of Earth's geographic North Pole.
\\
An early instrument designed to accomplish this task was the gyrocompass—a rapidly spinning disc mounted on a set of gimbals to induce gyroscopic precession. Nonetheless, its delicate mechanism exhibited a slow dynamic response to high frequencies, limiting its capacity to make rapid adjustments to changes in orientation. In spite of continuous advancements in manufacturing and miniaturization that have long replaced older technologies, the terminology has remained loyal to the historical name. Even today, regardless of the measurement technique employed, the task of finding the true North is widely recognized as \underline{gyrocompassing} \cite{groves2015principles}.
\\
Still, accurate determination of the heading angle remains a central navigation challenge. The extended 24-hour cycle generates an exceptionally faint signal with an amplitude as low as 0.00417 $^\circ$/sec, which weakens even further when moving away from the equator. Over time, technological breakthroughs have driven the pursuit of reducing its estimation error, attracting considerable attention from both academia and industry.
%
Various instruments have competed head to head, each presenting a unique detection mechanism, including measuring changes in capacitance, piezoelectric properties, resonance, vibratory modes, tuning forks, precession, or utilization of the Sagnac effect \cite{kourepenis1996performance, barbour2001inertial}. 
Yet, the lingering question remains: Have MEMS gyros reached the precision level required for tactical-grade applications? 
\\
With the ascent of machine learning (ML) models, the necessity for explicitly modeling intricate analytical errors has diminished, thanks to the ML's proficiency in capturing underlying patterns through learning \cite{el2020inertial, cohen2023inertial}.
\\
In this study, we introduce a deep learning framework that leverages the capabilities of an affordable MEMS sensor. By learning its temporal and spatial error patterns, our model demonstrates robustness to internal and environmental disturbances, allowing for accurate gyrocompassing within shorter time intervals. Our main contribution lies in three key aspects: 
\begin{enumerate}[label=(\roman*)]
    \item \textbf{Theory}: Development of learning framework allowing gyrocompassing for low-performance MEMS-based sensors. In addition, we introduce an angle-informed loss function as part of the proposed framework. 
    \item \textbf{Experimental Evaluations}: We show that our learning framework reduces waiting times by a factor of ten or, alternatively, cuts the alignment error by over 50\%. 
    \item \textbf{Open-Source Access}: To allow for results reproducibility and future research, all contents are shared @ \href{https://github.com/ANSFL/Learning-Based-MEMS-Gyrocompassing}{GitHub}.
\end{enumerate}

\noindent The remaining sections of the paper are structured as follows: Section \ref{sec:literature} reviews previous research endeavors, Section \ref{sec:theory} furnishes the theoretical background, and Section \ref{sec:PA} introduces the proposed approach. Section \ref{sec:AnR} presents and analyzes the results, while Section \ref{sec:disc} discusses them. Finally, Section \ref{sec:conc} brings the study to conclusions.

\section{Literature Review} \label{sec:literature}
As our study focuses on stationary self-alignment gyrocompassing, the primary challenge lies in achieving optimal estimates from the standalone INS. This section offers an exhaustive review of the notable works and methods, broadly classified into three thematic categories, each intended to extract the utmost information based on the specific use case. 

\subsection{Heuristic-based} \label{subs:heuristic}
Seminal works have laid the foundation for distinguishing between two key phases: an initial estimation stage, known as coarse alignment (CA), and an optional refinement stage, recognized as fine alignment (FA) \cite{britting2010inertial, chatfield1997fundamentals, titterton2004strapdown}. With the goal of promptly establishing a reasonable starting point, the choice between them hinges on the balance between the desired accuracy and the interference-free time window.
%
Park (1995) explored how correlation of the sensor bias and the cross-coupling terms can enhance the Euler angles observability, while Schimelevich investigated how inherent sensor noise affects the CA upper bound \cite{park1995covariance, schimelevich1996new}. 
\\
Jiang (1998) introduced comprehensive error analyses for strapdown INS (SINS) during on-ground self-alignment. Savage presented a two-speed orienting approach using direction cosine and quaternion forms, while in 2004, El-sheimy introduced a multi-level wavelet decomposition as a means to achieve noise-free CA \cite{jiang1998error, savage1998strapdown, el2004wavelet}. 
In 2012, Li introduced an innovative analytic calibration method aimed at mitigating gyro bias through a sensitivity analysis of Euler angles \cite{li2012error}. 
Liu, followed by Silva, leveraged geometric relationships among gravitational apparent vectors, enabling the extraction of the attitude matrix between the body and navigation frames based on motion vectors at three distinct time points and geographical data \cite{liu2014initial, silva2016error}.
In a more recent development (2020), Tian presented a MIMU/FOG compound system, featuring a cosine-fitting approach for calculating initial yaw angles, achieving superior performance with an accuracy of 0.1$^\circ$/hr \cite{tian2020cosine}.

\subsection{Filtering-based}
In principle, gathering more samples, whether through an increased sampling rate or additional sensors, should yield more useful information to reduce estimation errors. In practice, however, instrumental errors inevitably evolve over time, contaminating parameter estimates with non-systematic bias alongside increasing uncertainty \cite{unsal2012estimation}. 
%
To that end, filtering techniques, with the esteemed Kalman filter (KF) standing out prominently, tackle this challenge by employing a physical system model that relies on dynamic parameters, known as states, iteratively updated through a fusion mechanism. Upon receiving external readings, their significance is first assessed upon their error covariances, allowing for a constrained solution that aligns with the a-priori kinematics \cite{osman2006multi, guerrier2009improving, engelsman2023parametric}. 
\\
In their pioneering contributions, Lefferts (1982), followed by Grewal (1990), outlined the fundamental transition from parametric estimation to state estimation, employing the extended Kalman filter (EKF) to accommodate gyro and accelerometer error models \cite{lefferts1982kalman, grewal1990application}.
Scherzinger (1996) and Shin (2002), used non-holonomic constraints on stationary platforms in the form of zero velocity update (ZUPT) and zero integrated heading rate (ZIHR), providing reliable updates for both initialization and tracking tasks \cite{scherzinger1996inertial, shin2002accuracy}. 
\\
Chuanbin (2004), Wang (2009), and Acharya (2011) each introduced an enhanced self-alignment EKF scheme, employing augmented measurements to enhance azimuth misalignment angle accuracy during stationary initial alignment by establishing an update equation for estimating the east gyro drift rate \cite{chuanbin2004novel, wang2009fast, acharya2011improved}. 
%
In 2010, Iozan, and Collin, reported achieving 1-degree accuracy in determining true North orientation with a low-cost MEMS gyro, while accounting for small errors such as g-sensitivity and cross-axis coupling \cite{iozan2010north, iozan2010measuring}.
\\
Among other works over the past decade, Ali (2011) presented a second-order divided difference filter (DDF) \cite{ali2011initial}, Du (2016) introduced a disturbance observer-based Kalman filter (DOBKF) \cite{du2016fast}, and Klein (2018) demonstrated how kinematic constraints, when coupled with appropriate observability analysis, can effectively enhance unobserved states \cite{klein2018observability, klein2021ins, engelsman2022bias}.

\subsection{Optimization-based}
While the effectiveness of nonlinear filtering in tracking time-varying dynamics has never been questioned, their superiority under static conditions has been doubted by numerous optimization experts.
After in-field deployment, tactical platforms often remain stationary for extended periods before commencing their operational tasks. These nearly time-invariant conditions enable an abundance of uninterrupted measurements with low uncertainty and shared statistical properties. To maximize their utility, the unknown vector can be incorporated into an objective function, leveraging regression analysis to optimize the estimated angles \cite{bar1980azimuth, jiang1992error}.
\\
Arnaudov (2005), followed by Renkoski, pioneered the carouseling technique, wherein a leveled gyroscope undergoes continuous rotation ('slewing') around its z-axis. As the azimuthal distribution approaches uniformity, elements with opposite signs cancel each other out, minimizing both bias and noise presence \cite{arnaudov2005earth, renkoski2008effect}. 
\\
Johnson (2010) further refined it, achieving 2 milliradians (mrad) precision over a 5-minute integration period, and Prikhodko (2013), contrasted it with the two-point gyrocompassing technique ("maytagging") using single counterclockwise rotation \cite{johnson2010development, prikhodko2013mems}.
%
Other notable works in this category fall under the umbrella of optimization-based alignment (OBA) with Wu's (2012) heuristics to find the minimum eigenvector of the system, Li's (2014) fixed integral interval sliding method, and Chang extension to address the bias-attitude coupling \cite{chang2016optimization, wu2011optimization, li2014improved}. In a more recent study, Li (2018) introduced a novel gradient descent-based OBA, while Ouyang (2022) employed an analytically derived covariance error of the attitude angles \cite{li2018gradient, ouyang2022optimization}.
\\
As our survey reveals, despite numerous optimization techniques, there remains a gap in the literature when it comes to harnessing the transformative potential of learning-based approaches. It is this void we aim to fill in this paper.

\section{Theoretical Background} \label{sec:theory}
In this section, the fundamental principles that underpin our discussion are presented. Ranging from the central limit theorem to estimation theory, these concepts provide the foundation for understanding the complexities and performance boundaries that inertial sensors face.

\subsection{Inertial sensors model} 
Inertial sensors are electronic devices that convert continuous physical signals into evenly spaced discrete outputs, marked by $\tilde{{( \ )}}$. An accelerometer triad ($a$) measures specific forces (${\boldsymbol{f}}_{ib}^b$) experienced as 3D linear accelerations, while a gyroscope triad ($g$) records 3D angular velocities (${\boldsymbol{\omega}}_{ib}^b$). A commonly used linearized model describing the error relationship between the outputs and the ground truth (GT) input signals is as follows \cite{groves2015principles}:
\begin{align} 
\tilde{\boldsymbol{f}}_{ib}^b &= ( \textbf{I}_3 + \textbf{M}_a ){\boldsymbol{f}}_{ib}^b + {\textbf{\textit{b}}}_a + \textit{\textbf{w}}_a \ , \label{eq:f_ib} \\
\tilde{\boldsymbol{\omega}}_{ib}^b &= ( \textbf{I}_3 + \textbf{M}_g ) {\boldsymbol{\omega}}_{ib}^b + {\textbf{\textit{b}}}_g + \textit{\textbf{w}}_g \ \label{eq:w_ib} .
\end{align}
Here, $\textbf{I}_3$ represents a $3 \times 3$ identity matrix, and $\textbf{M}$ is a gain matrix to determine the error slope. Parameter $\textbf{\textit{b}}$ denotes the systematic bias, representing the sensor's accuracy, while $\textit{\textbf{w}}$ is modeled as a zero-mean white Gaussian noise, characterizing its inherent precision.
Under the reasonable assumption that the sensor is calibrated, both misprojection error ($\textbf{M} \approx 0$) and bias error ($\textbf{\textit{b}} \approx 0$) are considered negligible. 
\\
In principle, given known geographic coordinates, subtracting the bias-free outputs from their true counterparts results with a zero-mean white Gaussian distribution, with a zero off-diagonal covariance matrix ($\boldsymbol{\Sigma}$) describing its joint variability: 
\begin{align} 
\Delta {\boldsymbol{f}}_{ib}^b &= \tilde{\boldsymbol{f}}_{ib}^b - {\boldsymbol{f}}_{ib}^b \ \sim \ \mathcal{N}( \, \textbf{0}_3 \, , \boldsymbol{\Sigma}_a ) \ , \label{eq:res_acc} \\
\Delta {\boldsymbol{\omega}}_{ib}^b &= \tilde{\boldsymbol{\omega}}_{ib}^b - {\boldsymbol{\omega}}_{ib}^b \ \sim \ \mathcal{N}( \, \textbf{0}_3 \, ,  \boldsymbol{\Sigma}_g ) \ . \label{eq:res_gyr}
\end{align}
By exploiting the statistical independence of noise densities, variance along any of the $x,y,z$ axes can be reduced inversely proportional to the number of samples ($n$) through straightforward averaging over extended stationary periods
\begin{align} 
\sigma_{j,i}(n) = \sqrt{ \DIAG( \boldsymbol{\Sigma} ) / n } \ \ ; \
    \begin{array}{l}
      j \in \{a,g\} \\[1mm]
      i \in \{x,y,z\}
    \end{array} .
\end{align}
As long as stationarity is preserved, both the sample mean and sample variance are regarded as consistent estimators, as they steadily converge towards the underlying parameters. 
Being deterministic, yet unknown, the Cramér–Rao lower bound (CRLB) asserts that the highest precision of an unbiased estimator is the Fisher information \cite{bendat2011random, engelsman2023information}, given by
\begin{align} \label{eq:FIM}
\mathcal{I}(\mu) = - \EV \left[ \frac{\partial^2}{\partial \mu^2} \ell (\boldsymbol{x};\mu) \right] = ... = \frac{n}{\sigma^2} \ ,
\end{align}
where $\ell(x;\mu)$ is the log-likelihood function of sample $x$. Conversely, the reciprocal term places a lower bound for the variance error of estimator $\hat{\mu}$, as represented by
\begin{align} \label{eq:CRLB}
\VAR( \, \hat{\mu} \, ) \geq \frac{1}{ \mathcal{I}(\mu) } \ \Rightarrow \ \VAR( \, \hat{\mu} \, ) \geq \frac{\sigma^2}{n} \ .
\end{align}
However, the sworn enemies of that statement are revealed in the form of in-run variations. 
The well-known bias instability (BI), signifies the slow, time-varying drift of the sample mean away from the GT input value when the system is at rest. While accelerometers have BI specified in mg, gyros typically use a one-hour time frame $^\circ$/hr, allowing for a comparative perspective with Earth's rotation rate (15 $^\circ$/hr).
\\
The next error parameter characterizes the sensors' in-run variability based on their integrated quantities. The accumulation of the underlying noise induces variability, expressed as velocity random walk (VRW) in accelerometers, measured in m/s/$\sqrt{\text{hr}}$, and angle random walk (ARW) in gyroscopes, measured in $^\circ$/$\sqrt{\text{hr}}$. Effectively, these parameters quantify the temporal increase in standard deviation, and can be recovered by multiplying them by the square root of the time.
\\
While these factors offer a factual assessment of the sensor's quality, they also play a crucial role in defining the statistical limitations of the estimator's efficiency. 
In-run variations disrupt the normality assumption of the stationary distribution, prompting the quest for innovative, non-model-based approaches that can effectively account for these variations while preserving empirical superiority.


\subsection{Self-alignment model} 
\begin{figure}[h] 
\begin{center}
\includegraphics[width=0.32\textwidth]{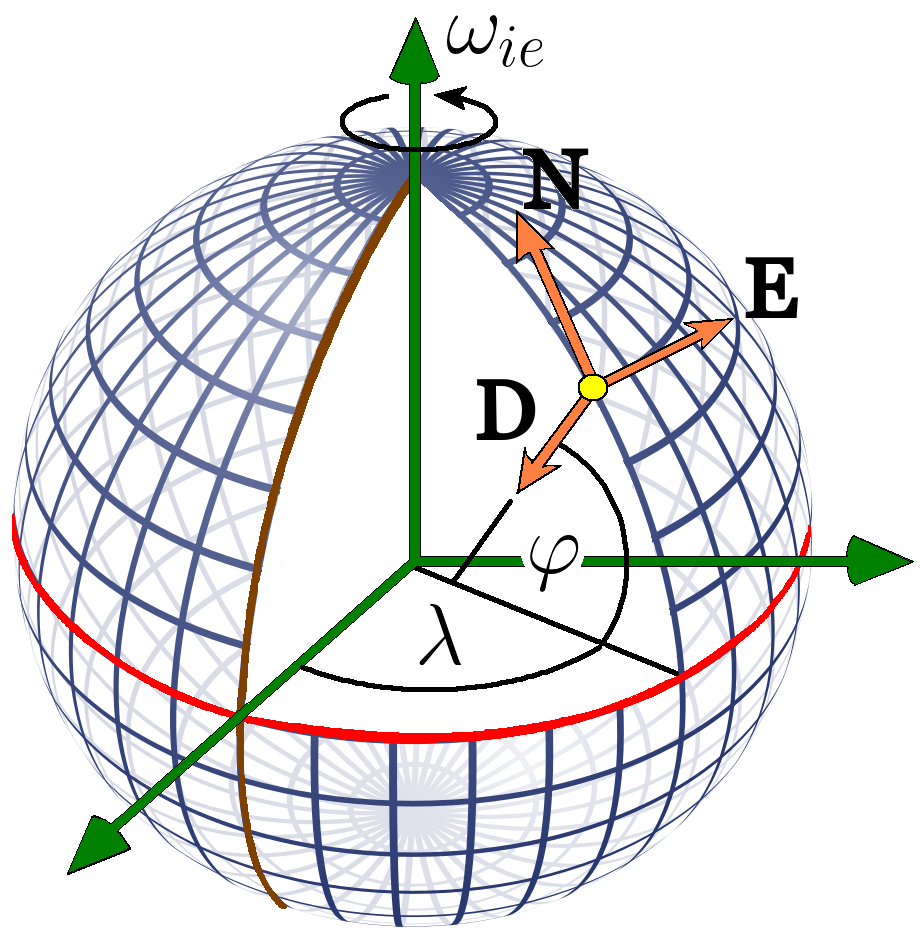}
\caption{Diagrammatic representation of ECEF and NED frame.}
\label{fig:Sphere}
\end{center}
\end{figure}

Once stationarity is detected ($ \| \boldsymbol{f}_{ib}^b \| \approx \| \textbf{g}^n \|$), the standalone alignment seeks to establish orientation using onboard sensors only. As inertial sensors quantify the dynamics acting within the body frame, it is essential to project them onto a local inertial frame, free from Coriolis and centrifugal forces induced by Earth's rotation.
To that end, an orthogonal transformation matrix is used to describe rotations between frames, such as from the body frame ($b$) to the navigation frame ($n$), and vice versa with $\textbf{T}_n^b = (\textbf{T}_b^n)^{\TT}$. 
\\
By applying three consecutive Euler angle rotations (yaw $\psi$, pitch $\theta$, and roll $\phi$) about the z-y-x axes respectively, the transition achieves seamless alignment with Earth's North, East, Down (NED) system as follows: 
\begin{equation} \label{eq:T_b_n}
\textbf{T}_{b}^n = \begin{bmatrix}
\text{c}_\theta  \text{c}_\psi  &  \text{s}_\phi \text{s}_\theta \text{c}_\psi - \text{c}_\phi \text{s}_\psi  &   \text{c}_\phi \text{s}_\theta \text{c}_\psi + \text{s}_\phi \text{s}_\psi \\
\text{c}_\theta \text{s}_\psi & \text{s}_\phi \text{s}_\theta \text{s}_\psi +  \text{c}_\phi \text{c}_\psi & \text{c}_\phi \text{s}_\theta \text{s}_\psi - \text{s}_\phi \text{c}_\psi  \\
-\text{s}_\theta & \text{s}_\phi \text{c}_\theta & \text{c}_\phi \text{c}_\theta 
\end{bmatrix} \ ,
\end{equation}
\\
where 's' and 'c' represent the sine and cosine functions. 
\\
The initial alignment step, referred to as 'leveling', determines the attitude angles ($\phi, \theta$) between the local horizontal plane and the local vertical. Being stationary, the only specific force sensed by the accelerometers is the gravity reaction in the negative down direction, as given by
\begin{align} \label{eq:gravity}
\boldsymbol{f}_{ib}^b = 
\begin{bmatrix} 
\ f_{ib, x}^b  \ \ \\ \ f_{ib, y}^b  \ \ \\ \ f_{ib, z}^b \ \ 
\end{bmatrix} = -\textbf{T}_n^b \, \textbf{g}^n  = \begin{bmatrix} 
s_{\theta} \\ - s_{\phi} c_{\theta}  \\ - c_{\phi} c_{\theta} 
\end{bmatrix} \text{g} \ .
\end{align}
With only two unknowns, this overdetermined system produces the following gravity-independent solution \cite{groves2015principles, li2012error}:
\begin{align}
\phi &= \arctan_2 \Big( -f_{ib, y}^b \, , \, -f_{ib, z}^b \Big) \ , \label{eq:roll} \\
\theta &= \arctan \Big( \frac{ f_{ib, x}^b }{ \sqrt{ f_{ib, y}^{b \, 2} + f_{ib, z}^{b \, 2} } } \Big) \ . \label{eq:pitch}
\end{align}
After achieving proper leveling, the third Euler angle ($\psi$) can be determined by extracting the rotation components about the z-axis of an Earth-fixed coordinate frame (ECEF), in a process known as \underline{gyrocompassing}. The stationary gyros are than expected to output $\| \boldsymbol{\omega}_{ib}^b \| \approx \omega_{ie}$, as described by 
\begin{align}
\boldsymbol{\omega}_{ib}^b = \textbf{T}_{n}^b \textbf{T}_{e}^n
\boldsymbol{\omega}_{ie}^e \ .
\end{align}
whereas the mapping from the ECEF frame (green) onto the North-East-Down (NED) frame (orange) is given by $\textbf{T}_{e}^n$, succeeded by projecting onto the sensor frame (yellow circle), as illustrated in Fig.~\ref{fig:Sphere}. 
\\
Longitudes ($\lambda$) define east-west positioning relative to the prime meridian (brown), latitudes ($\varphi$) indicate north-south positioning in relation to the equator (red), and the down component (\textbf{D}) points approximately toward Earth's center.
Since the local NED frame is inherently tangential to the Earth's surface, rotation is devoid of an eastward component, projecting exclusively onto the north and downward directions: 
\begin{align}
\textbf{T}_{b}^n \boldsymbol{\omega}_{ib}^b = 
\begin{bmatrix} 
-s_\varphi c_\lambda & - s_\varphi s_\lambda & c_\varphi \\
-s_\lambda & c_\lambda & 0 \\
-c_\varphi c_\lambda & - c_\varphi s_\lambda & -s_\varphi
\end{bmatrix} \begin{bmatrix} 0 \\ 0 \\ \omega_{ie} \end{bmatrix} = 
\begin{bmatrix} c_\varphi \\ 0 \\ -s_\varphi \end{bmatrix} \omega_{ie} \ .
\end{align}
Either here, the resulting system does not necessitate prior knowledge of the position; instead, it exclusively derives the heading angle from the computed attitude angles
\begin{align}
\text{s}_{\psi} &= -\omega_{ib,y}^b \text{c}_{\phi} + \omega_{ib,z}^b \text{s}_{\phi} \ , \\
\text{c}_{\psi} &= \ \omega_{ib,x}^b \text{c}_{\theta} + \omega_{ib,y}^b \text{s}_{\phi} \text{s}_{\theta} + \omega_{ib,z}^b \text{c}_{\phi} \text{s}_{\theta} \ ,
\end{align}
solvable through the following four-quadrant arctangent \cite{titterton2004strapdown}
\begin{align} \label{eq:psi_GC}
\underline{\psi} &= \arctan_2 \big( \text{s}_{\psi} \, , \, \text{c}_{\psi} \big) \ , 
\end{align}
and after leveling, can be simplified into 
\begin{align} \label{eq:psi_GC_simp}
\underline{\psi} = \arctan_2 \big( -\omega_{ib,y}^b \, , \, \omega_{ib,x}^b \big) \ .
\end{align}
For example, when conducting our experiment at $\varphi$=32.76$^\circ$N, we observed a maximum signal strength of
\begin{align}
\| ( \omega_x, \omega_y ) \| = \text{c}_{\varphi} \omega_{ie} \approx 0.0035 \ \text{[deg/s]} \ ,
\end{align}
indicating a 16\% decay in planar components due to our location being 3,500 km north of the equator.

\subsection{First-order error analysis}
The closed-form nature of the Euler angles enables further examination through linear error analysis, offering additional insights into their sensitivity to the input parameters by using
\begin{align} \label{eq:analysis}
\Delta \, \textit{f} \, (\textit{x}_i, ..., \textit{x}_n ) = \vec \nabla \, \textit{f} \cdot \Delta \textit{\textbf{x}} = \sum_i^n \left( \frac{\partial \textit{f}}{\partial \textit{x}_i} \Delta \textit{x}_i \right) \ .
\end{align}
To simplify notation, the inertial-to-body frame (ib) subscript is omitted from both sensor notations (\ref{eq:res_acc},\ref{eq:res_gyr}) henceforth. Applying \eqref{eq:analysis} on \eqref{eq:roll}, the roll angle error is given by
\begin{align}
\Delta \phi (\boldsymbol{f}^b) = \vec{\nabla} \phi \cdot \Delta \boldsymbol{f}^b = \frac{\partial \phi}{\partial f_{y}^b} \, \Delta {f}_{y}^b + \frac{\partial \phi}{\partial f_{z}^b} \, \Delta {f}_{z}^b \ .
\end{align}
In a near-horizontal condition, the horizontal x-y forces are minuscule compared to the vertical gravity by several orders of magnitude (OoM) and can therefore be disregarded
\begin{align}
\Delta \phi = \frac{ f_z^b \, \Delta f_y^b - \cancelto{0}{f_y^b} \, \Delta f_z^b }{ \cancelto{0}{f_y^{b \, 2}} + {f}_{z}^{b \, 2 } } \approx \frac{ \Delta f_y^b }{ \text{g} } \ .
\end{align}
In a similar manner, applying \eqref{eq:analysis} on \eqref{eq:pitch}, followed by assuming $f_x^b = f_y^b \approx 0$, yields the following pitch angle error
\begin{align} \label{eq:err_roll}
\Delta \theta (\boldsymbol{f}^b) = \vec{\nabla} \theta \cdot \Delta {\boldsymbol{f}}^b = \frac{\partial \theta}{\partial f_{x}^b} \, \Delta {f}_{x}^b + \frac{\partial \theta}{\partial f_{y}^b} \, \Delta {f}_{y}^b + \frac{\partial \theta}{\partial f_{z}^b} \, \Delta {f}_{z}^b \notag \\ 
\ = \frac{ \left( f_{y}^{b \, 2} + f_{z}^{b \, 2} \right) \, \Delta  f_{x}^{b} - f_x^b f_y^b \, \Delta f_y^b + f_x^b f_z^b \, \Delta f_z^b }{ {\| {\boldsymbol{f}}^b \|}^2 \sqrt{ f_{y}^{b \, 2} + f_{z}^{b \, 2} } } \approx  \frac{ \Delta  f_{x}^{b} }{ \text{g} } \ . 
\end{align}
Regarding the error of the heading angle \eqref{eq:psi_GC}, its determination relies not only on gyroscope measurements, but also indirectly on the accelerometers through the attitude angles:
\begin{align} \label{eq:err_pitch}
\Delta \psi (\boldsymbol{f}^b, \boldsymbol{\omega}^b) &= \vec{\nabla} \psi \cdot \Delta \boldsymbol{f}^b + \vec{\nabla} \psi \cdot \Delta \boldsymbol{\omega}^b = 
\frac{\partial \psi}{\partial f_{x}^b} \Delta f_{x}^b 
\\
 +\frac{\partial \psi}{\partial f_{y}^b} \Delta f_{y}^b &+ \frac{\partial \psi}{\partial f_{z}^b} \Delta f_{z}^b
+ \frac{\partial \psi}{\partial \omega_{x}^b} \Delta \omega_{x}^b + \frac{\partial \psi}{\partial \omega_{y}^b} \Delta \omega_{y}^b  + \frac{\partial \psi}{\partial \omega_{z}^b} \Delta \omega_{z}^b \ . \notag
\end{align}
However, due to the intricate and extensive derivation involved, only the final solution is presented herein
\begin{align} \label{eq:err_psi}
\Delta \psi (\boldsymbol{f}^b, \boldsymbol{\omega}^b) \approx - \frac{ \Delta f_{y}^b }{ \text{g} } \tan \varphi + \frac{ \Delta \omega_{y}^b }{ \omega_{ie} } \sec \varphi \ ,
\end{align}
revealing a latitude-sensitive interplay between both sensor errors, as discussed in details in \cite{groves2015principles, britting2010inertial}. 
%
Nevertheless, the disparity between the leveling error and the gyrocompassing error merits careful attention:
Both the roll \eqref{eq:err_roll} and the pitch errors \eqref{eq:err_pitch} enjoy a pronounced gravity signal, which diminishes the attitude errors by a factor of 1/$\text{g}$. 
\\
For example, a mid-tier MEMS accelerometer with a 1 milli-g (mg) resolution can achieve an outstanding attitude accuracy within a range of 1 mrad. 
Conversely, a MEMS gyroscope of the same grade, with a commendable precision of 0.5 mrad/sec (or 100 $^\circ$/hr), would fall far short of reaching an adequate heading angle with just a single measurement.
\\
As a two-variable function, the heading error \eqref{eq:err_psi}  
is comprised of two of subcomponents; a gravity term, which appears to be insignificant away from the poles, and an Earth rotation term that holds sway over the entire expression. 
At any latitude, an angular velocity of 73 µrad/sec (or 15 $^\circ$/hr) in the denominator markedly amplifies gyro errors in the numerator. Thereby, even the slightest disturbance has the potential to overpower the delicate rotational signal, which is several OoMs smaller.

\section{Proposed methodology} \label{sec:PA} 
After establishing the theoretical background and describing the gyrocompassing challenge resulting from the weak rotation signal, in this section, our approach is presented with the aim of achieving a superior performance benchmark.

\subsection{A learning approach}  
\begin{figure}[t] 
\begin{center}
\includegraphics[width=0.5\textwidth]{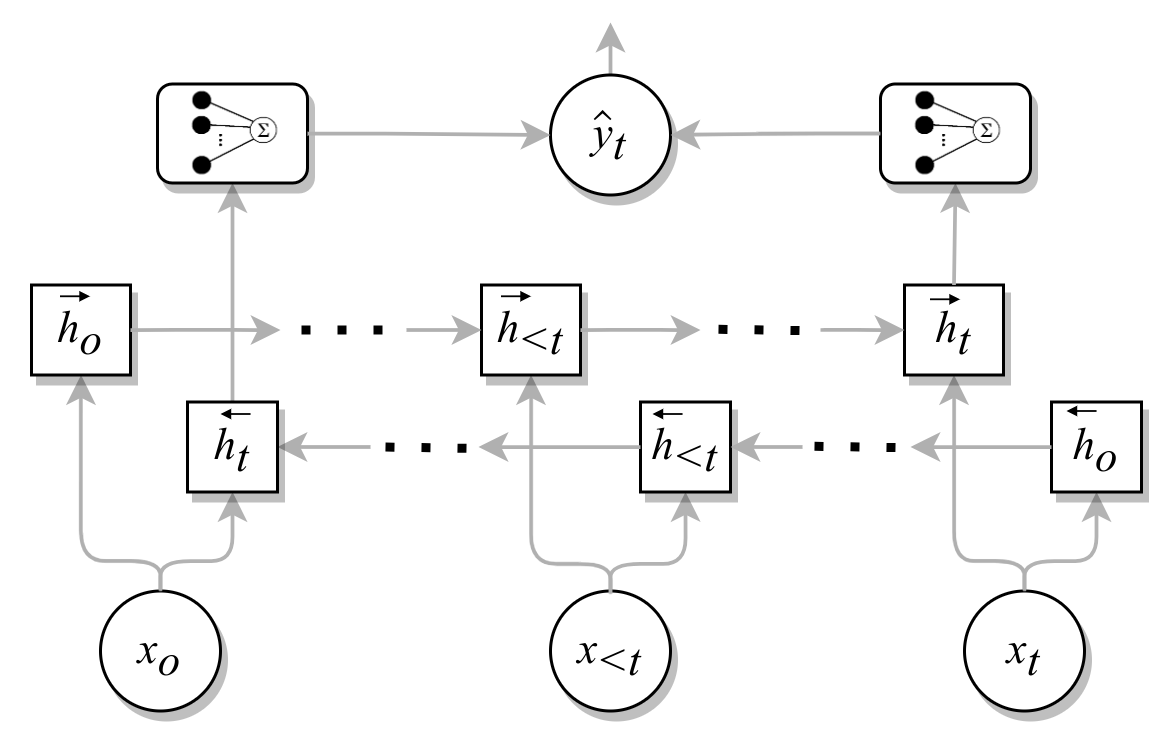}
\caption{Unrolled computational graph illustrating our Bi-directional LSTM model performing a many-to-one mapping.}
\label{fig:BiLSTM}
\end{center}
\end{figure}

As indicated in \eqref{eq:CRLB}, the highest achievable performance, or conversely, the smallest attainable error of an unbiased estimator, is either upper or lower bounded (respectively) by the number of stationary samples available. However as stochastic errors begin to manifest within the sensor stream, the limitations of relying on the sample mean for obtaining maximum likelihood estimates (MLE) become apparent. 
\\
To address the challenges posed by emerging errors, we propose a learning model in lieu of conventional model-based approaches. Leveraging the noise reduction capabilities and adeptness at handling nonlinear behavior, neural networks are particularly effective for mitigating the problem at hand. 
\\
Opting for a recurrent neural network (RNN) architecture is natural due to its inherent capacity to learn complex dependencies that may become apparent over long-term intervals. Fig.~\ref{fig:BiLSTM} illustrates our empirically chosen model, which excels in extracting meaningful patterns apart from the random ones. The gyro readings at the base layer, denoted by $x_0, ..., x_t$, are fed into hidden cells marked by $h_t$, forming a bi-directional long short-term memory model (Bi-LSTM). 
\\
Each time step is processed both in the positive time direction, $\overrightarrow{h_t}$, and in the negative time direction (anti-chronologically), denoted as $\overleftarrow{h_t}$. 
\\
To reduce dimensionality and enhance temporal representations, the processed sequences from both directions are ultimately directed to an individual fully connected layer, facilitating a many-to-one mapping with an unbounded output range. Upon completion, the dense outputs are combined via element-wise summation to generate the model's prediction, namely the estimated heading angle, denoted as $\hat{y}$.

\subsection{Loss function}
Unlike linear scales which span values along a unidirectional sequence, the cyclic nature of angles necessitates a continuous error across all four quadrants, capped at 180$^\circ$. To address this constraint, we introduce an angle-informed loss function detailed in Algorithm~\ref{alg:cmse}, under the title of cyclic mean squared error (CMSE). 
\\
In this process, error batches between ground truth labels ($y$) and model predictions ($\hat{y}$) undergo a suitable transformation, ensuring quadrant-aware continuity in compliance with the azimuthal circle. This streamlines the computation of gradients from batch errors, facilitating the seamless updating of model parameters in the training process.
\begin{algorithm}[h] 
\caption{Cyclic mean squared error}
\label{alg:cmse}
\SetAlgoLined
\SetKwFunction{dToR}{Deg2rad}
\SetKwFunction{rToD}{Rad2Deg}
\SetKwFunction{CyclicMSELoss}{CyclicMSELoss}
\SetKwFunction{atanTwo}{atan2}
\SetKwFunction{sin}{sin}
\SetKwFunction{cos}{cos}
\SetKwFunction{mean}{mean}
\SetKwProg{Fn}{Function}{:}{}
\Fn{\CyclicMSELoss{$y$, $\hat{y}$}}{ 
    $y, \, \hat{y} \leftarrow \dToR(y), \, \dToR(\hat{y})$ \\
    $\Delta y \leftarrow y - \hat{y}$ \\
    $err \leftarrow \atanTwo( \, \sin( \Delta y ) \, , \, \cos( \Delta y ) \, ) $ \\
    $\mathcal{L}_{\text{CMSE}} \leftarrow \mean \Big( \big(  \rToD(err) \big)^2 \Big)$ \\ 
    \KwRet $\mathcal{L}_{\text{CMSE}} \hspace{2mm}$ \tcp{[deg$^2$]}
    }
\end{algorithm}

\subsection{Data augmentation}
Despite many hours of measurements, deep learning models demand a substantial volume of samples to facilitate seamless interpolation within their output range. 
To address this, data augmentation becomes a crucial strategy, artificially expanding the training dataset through various transformations and perturbations, enhancing performances on diverse inputs. 
By exploiting the stationarity of the problem, both data points and their labels can be perturbed using several linear operators such as rotational transform ($\texttt{rotate}$), additive white Gaussian noise ($\texttt{AWGN}$), and random bias addition ($b_i$), all while preserving their structural integrity, as detailed in Algorithm~\ref{alg:aug}. 

\begin{algorithm}[h]
\caption{Data augmentation algorithm} 
\label{alg:aug}
\SetAlgoLined
\SetKwInOut{Input}{Input}
\SetKwInOut{Output}{Output}
\Input{orinigal data ( $\mathcal{X} , \mathcal{Y}$ )}
%
\SetKwFunction{Tensor}{torch.tensor}
\SetKwFunction{FMain}{dataAugment}
{
    $\widetilde{\mathcal{X}} , \widetilde{\mathcal{Y}}$ $\leftarrow$ [~]\,, [~] \tcp*{Initialization}
    $\Psi \leftarrow \psi_{min} : \Delta \psi : \psi_{max}$\\
    \For{$(i \, , \, x_i)$ in $\mathcal{X}$}{
        \For{$ \psi_i$ in $\Psi$}{
            $temp \leftarrow$ \texttt{Rotate}($x_i$, $\textit{angle}=\psi_i$) \\
            $temp \leftarrow$ \texttt{AWGN}( $temp$ ) + $b_i$ \\ 
            $\widetilde{\mathcal{X}}\texttt{.add}$( $temp$ ) \\
            $\widetilde{\mathcal{Y}}\texttt{.add}$( $\mathcal{Y} \, [\,i\,] + \psi_i$ )\\
        }
    }
}
\KwRet \texttt{tensor}( $\widetilde{\mathcal{X}} , \widetilde{\mathcal{Y}}$ ) \\
\end{algorithm}

\section{Experimental Results} \label{sec:AnR}
In this section we showcase the experimental settings, present the outcomes, assess their alignment with relevant literature, and analyze the performance of our model in comparison to that of the conventional baseline.

\begin{figure}[b]
\begin{center}
\includegraphics[width=0.49\textwidth]{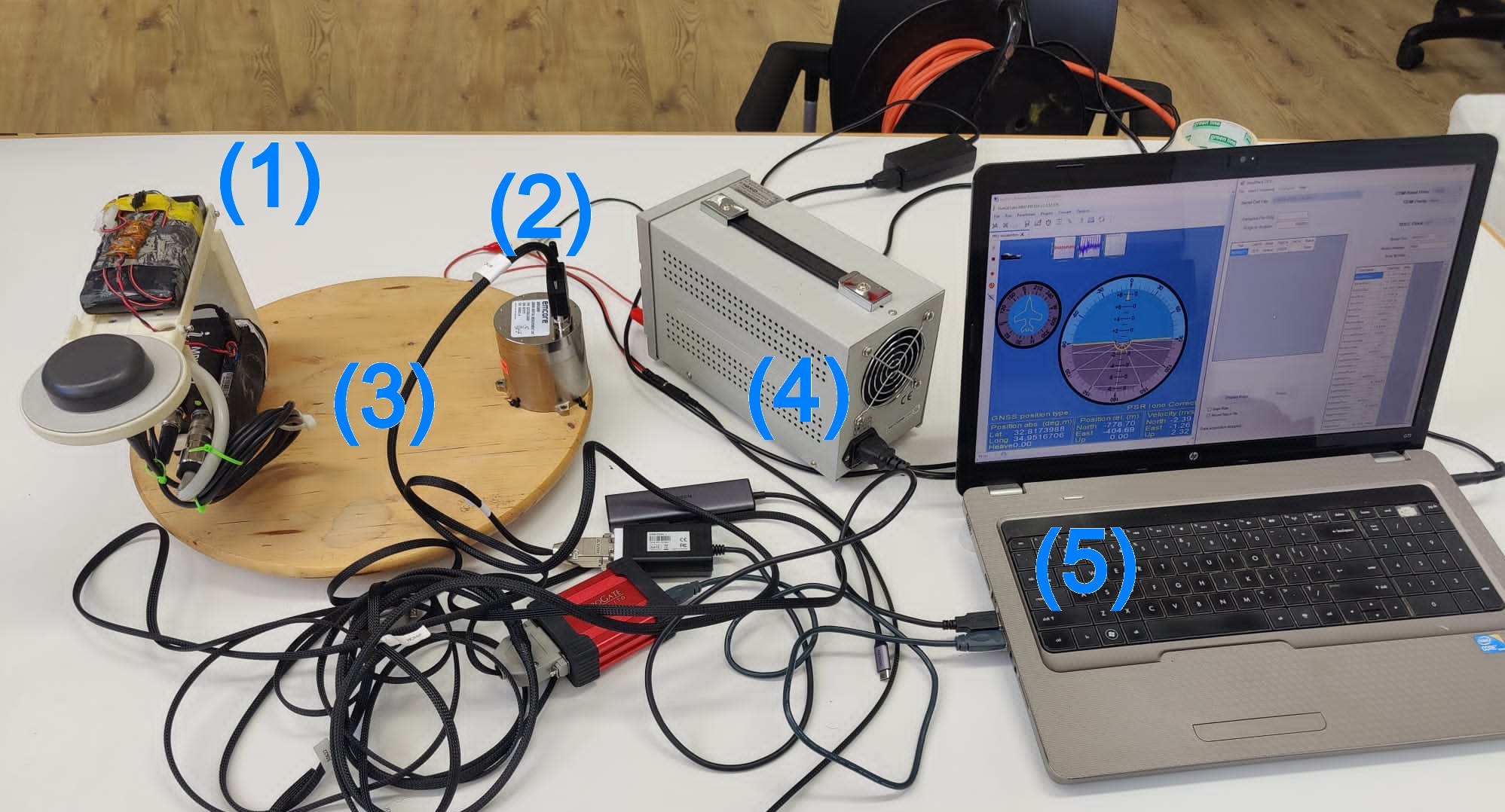}
\caption{Layout of the experimental setup.}
\label{fig:setup}
\end{center}
\end{figure}
\subsection{Experimental setup}
\begin{figure}[h]
\begin{center}
\includegraphics[width=0.5\textwidth]{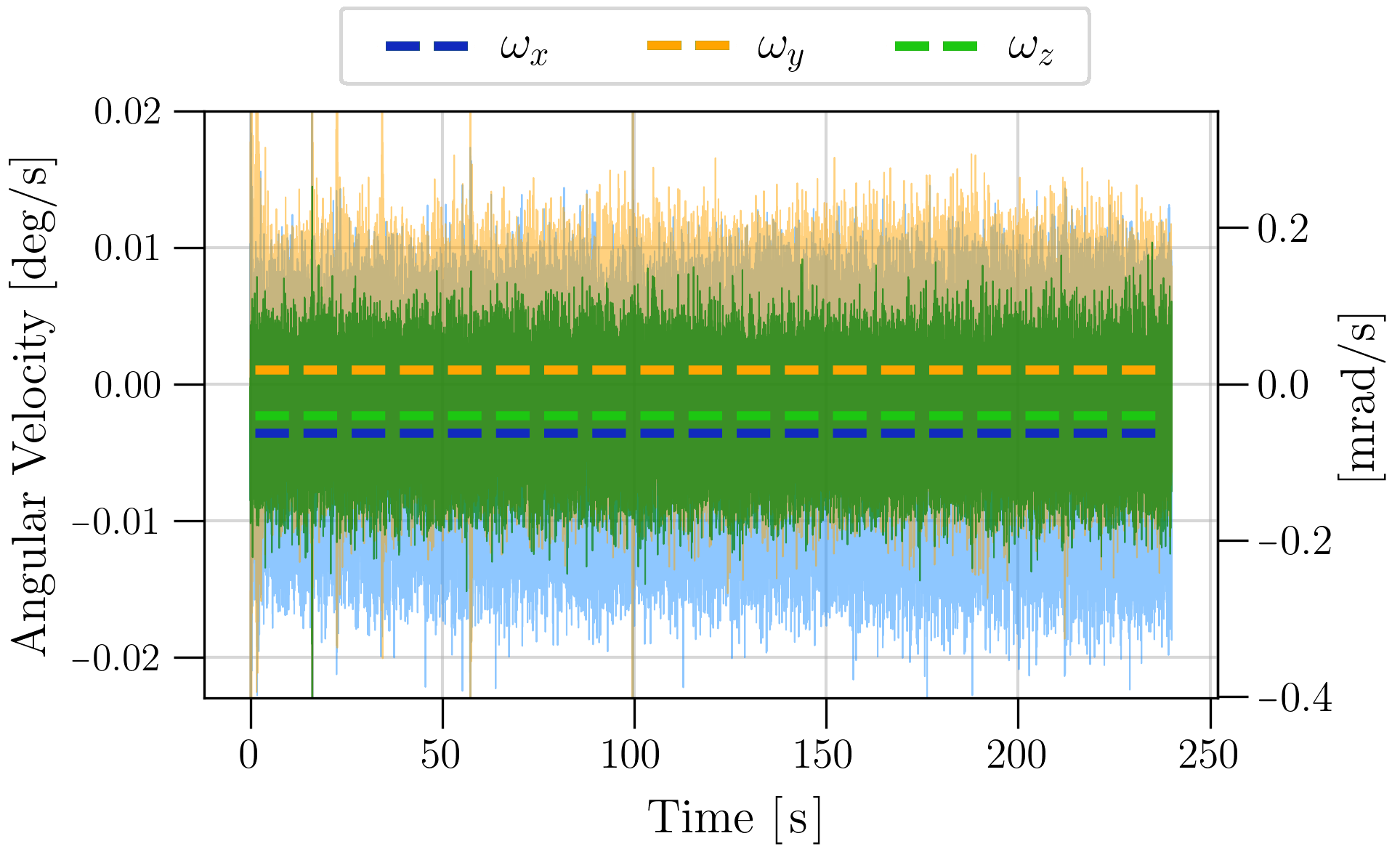}
\caption{An illustrative example of a 3D gyro measurements.}
\label{fig:measurements}
\end{center}
\end{figure}

To ensure robust generalizability, it is crucial to acquire data of both quantity and quality. Figure~\ref{fig:setup} illustrates our experimental setup conducted under controlled laboratory conditions, highlighting its five main components within blue parentheses:
\begin{enumerate}
   \item Control module\footnote{MRU-P datasheet @ \url{https://www.inertiallabs.com/mru-datasheet}}: Ensures level conditions and
   provides GT heading angles ($y$) with a stationary accuracy of $0.2^\circ$.
   \item Test module\footnote{Emcore SDC500 datasheet @ \href{https://emcore.com/wp-content/uploads/2022/05/966762_B-SDC500.pdf}{https://emcore.com/SDC500.pdf}}: Positioned at the opposite end of the diameter, our MEMS-IMU provides stationary measurements ($x_0, ..., x_t$) at an opposing heading angle ($y-180^\circ$), with specified BI of 1$^\circ$/hr and ARW of 0.02$^\circ$/$\sqrt{\text{hr}}$.
   \item Rotating plate: Both sensors are positioned on a levelled plate that rotates freely around its azimuth axis, allowing stationary measurements across various heading angles.
   \item Power supply: Ensures a stable and reliable source of energy for uninterrupted system functionality.
   \item Computing unit: Serves as the central processing hub, facilitating efficient operation and ensuring that real-time data is saved, labeled, and appropriately organized.
\end{enumerate}
While in stationary conditions, measurements were sampled at 600 Hz for 4 minutes, maintaining an average interval of 5$^\circ$ across the entire 360$^\circ$ azimuth plane. The overall dataset comprises 80 samples, randomly split into training, validation, and test sets with a ratio of 70:10:20, where training was conducted using a single Nvidia T4 GPU. 
\\
Fig.~\ref{fig:measurements} provides a representative visualization of a random gyroscope sample lasting 4-minute across all three 3 channels, resulting in a dimensionality of 144,000$\times$3.
While the dense measurements form the colored background for each axis, their corresponding sample means are marked with dashed lines to emphasize the specific axial projection of $\omega_{ie}$.

\subsection{Sensor performances}
\begin{figure}[t] 
\begin{center}
\includegraphics[width=0.495\textwidth]{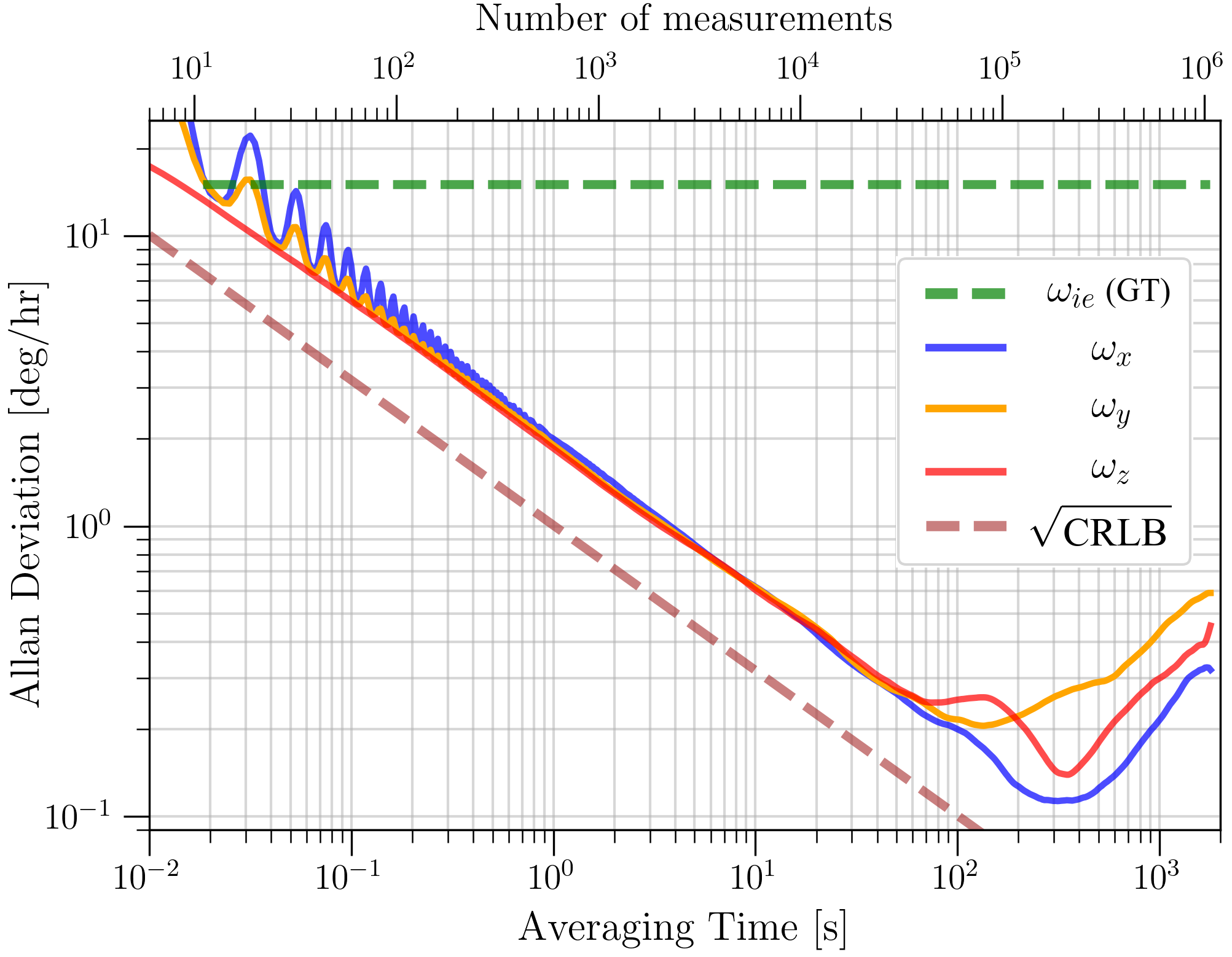}
\caption{Allan deviation analysis for the gyroscope sensor data.}
\label{fig:AV}
\end{center}
\end{figure}

\begin{figure}[t] 
\begin{center}
\includegraphics[width=0.5\textwidth]{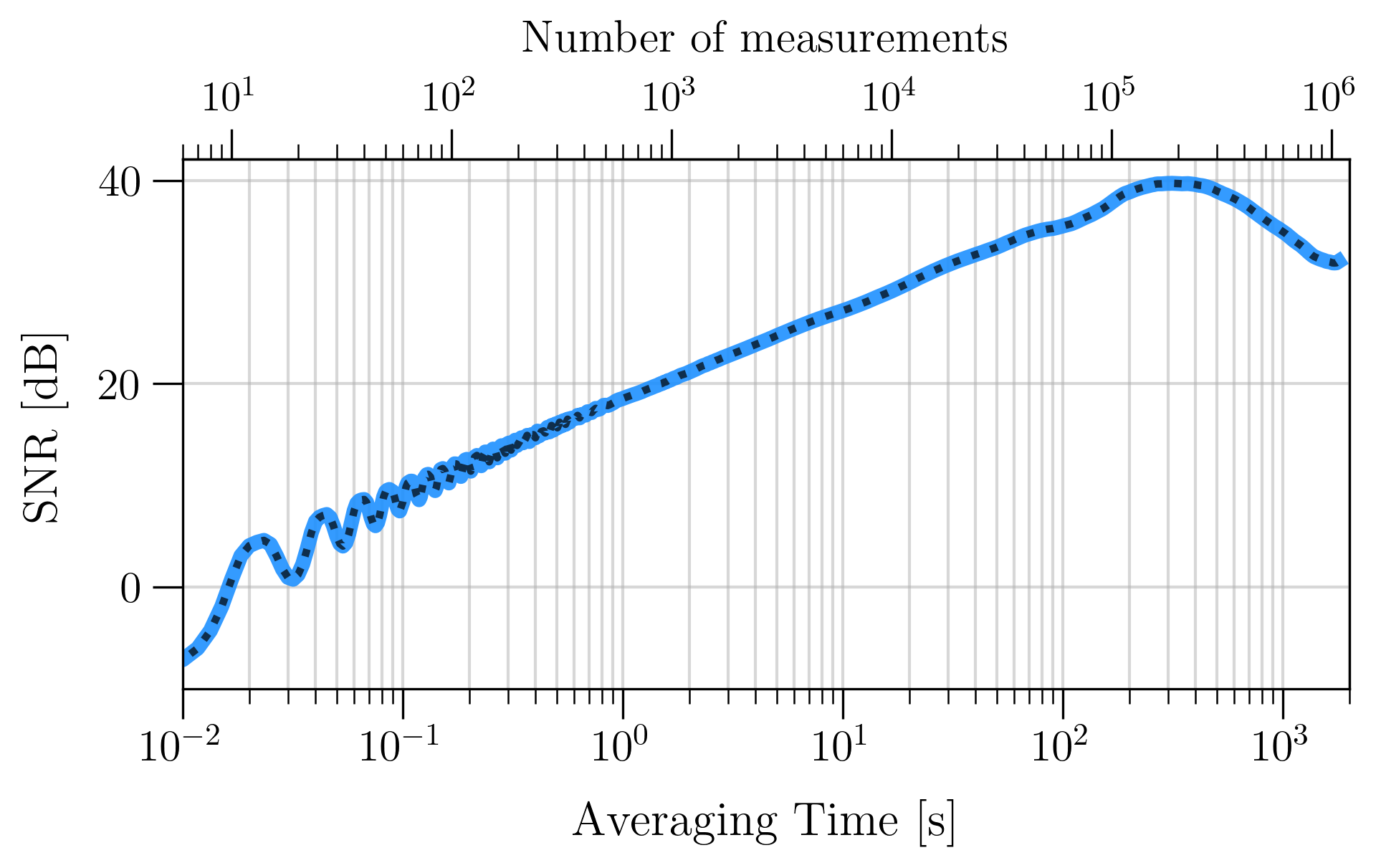}
\caption{SNR of the sample mean vs averaging time.}
\label{fig:err_SNR}
\end{center}
\end{figure}

Prior to evaluating the model performances, an empirical verification of sensor parameters is undertaken to ensure their alignment with the current experimental day. Under steady-state conditions, a 1-hour static measurement is taken using the Allan Variance (AV) analysis \cite{allan1966statistics}. This disturbance-free scenario aims to isolate error variables and quantify their dynamics through extensive time averaging, as given in
\begin{align} \label{eq:avar}
\sigma_x^2(\tau) = \frac{1}{2(n-2\tau)} \sum_{i=0}^{n-2\tau-1} \big( x_{i+2\tau} - 2x_{i+\tau} + x_{i} \big)^2 \, .
\end{align}
Here, $n$ represents the number of stationary measurements, imposing a computational complexity of $\mathcal{O}(n)$, while the time lag $\tau$ signifies the observation period. 
\\
As a translation-invariant operator, it remains unaffected by bias remainder or incomplete calibration, allowing a focus on the long-term stability of the output stream.
\\
The square root of the AV analysis \eqref{eq:avar}, wherein the asymptotic relationships of noise densities are evaluated against the number of averaged measurements using is presented in Fig.~\ref{fig:AV}.
\\
To straighten their power-law dependence over time, a base-10 log scale is employed, facilitating the placement of all three axial estimates within an intuitive error range: the upper tolerance is determined by the Earth rotation rate (green), while the lower one is indicated by the Cramér–Rao lower bound (brown), by setting its specified noise density in \eqref{eq:CRLB}. 
As can be seen, the estimates can be roughly classified into three distinct characteristic error regimes and compared with the manufacturer specifications \cite{el2007analysis}: 
\begin{enumerate}[label=(\roman*)]
    \item Angle random walk ($t \lesssim$ 250 s): Random accumulation of integrated gyro signals induced by the approximate white noise spectrum over short time scales. This results in a reduction rate of one OoM over a span of two decades (dec) of time, demonstrating a negative slope of -0.5. Here, the empirical ARW becomes roughly 0.03$^\circ$/$\sqrt{\text{hr}}$.
    \item Bias instability ($t \approx$ 250 s): Determines the maximum averaging time beyond which the error ceases to decrease, transitioning to a flat slope of 0. Following proper scaling, the BI can be recovered from the minima of the graphs, here around 0.2$^\circ$/hr, effectively explaining the expected drift over the course of one hour.
    \item Rate random walk ($t \gtrsim$ 250 s): Indicates low-frequency drift resulting from prolonged sensor use, which remains immune to averaging. Its non-stationary nature causes errors to grow proportionally to $\sqrt{t}$ with a logarithmic slope of +0.5. This aspect is particularly relevant for significantly longer measurements.
    \end{enumerate}
Another important insight evident from this graph is the efficiency limitation imposed by time. Interestingly, during short periods where noise steadily decreases, the sample means are slightly above the error bound ($\sqrt{\text{CRLB}}$). However, upon reaching the BI accumulation threshold, the non-stationary effects introduce uncompensated bias that gradually diminishes the efficiency of the estimates.
\\
Fig.~\ref{fig:err_SNR} emphasizes the pivotal role of the averaging time in signal refinement, using the conventional signal-to-noise ratio (SNR) formula, with $\omega_{ie}$ in the numerator and the reduced noise in the denominator.
%
\begin{figure}[b] 
\begin{center}
\includegraphics[width=0.49\textwidth]{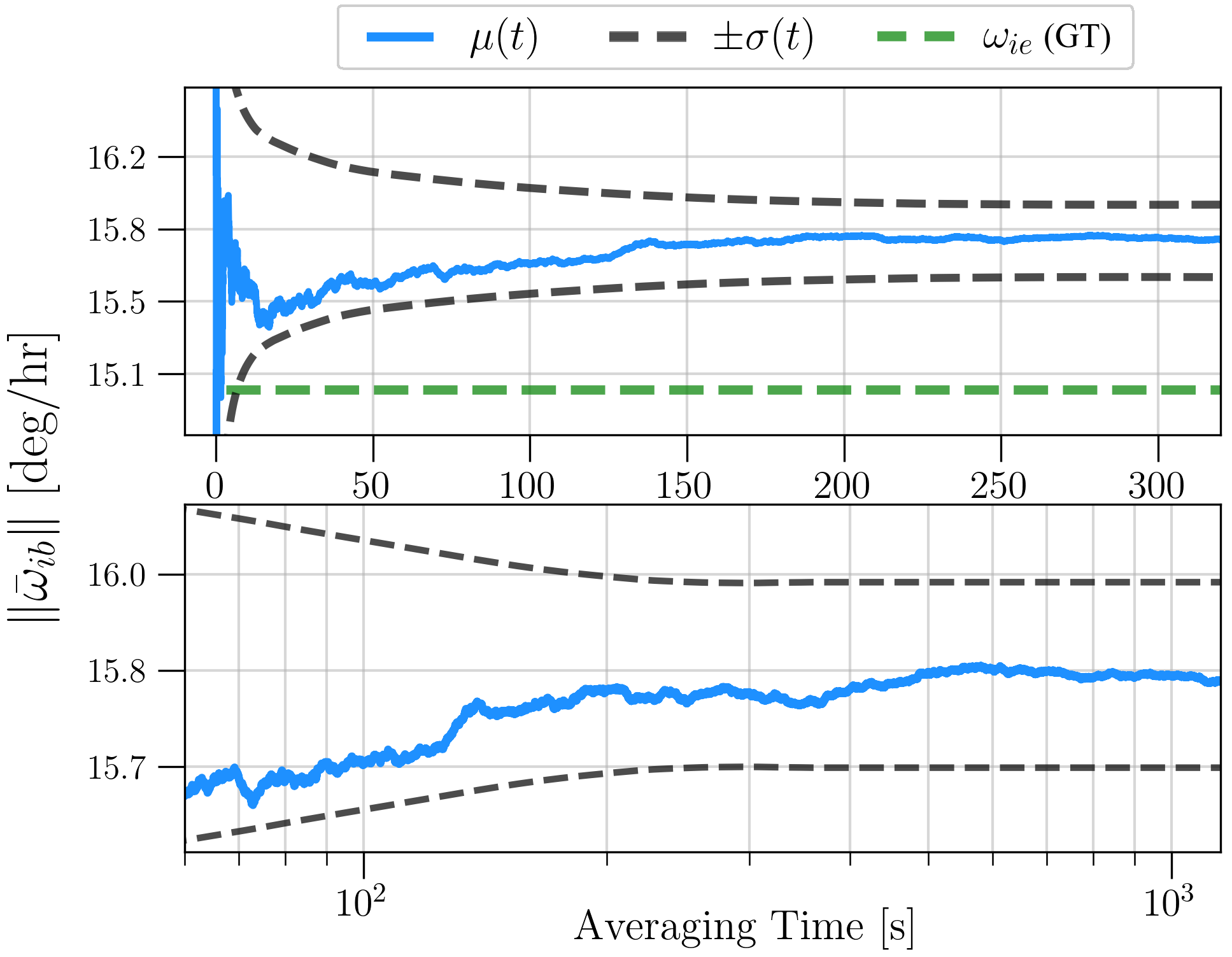}
\caption{Sensitivity of the sample mean to Earth rotation ($\omega_{ie}$).} 
\label{fig:err_ma}
\end{center}
\end{figure}
At the outset, a negative SNR is observed, signifying that within a small number of samples, sensor noise dominates the delicate rotation signal. Then, with an increasing number of samples being averaged, the signal \underline{amplitude} gradually intensifies at an approximate rate of 1 OoM (20 dB) over two decades (dec), or about 10 dB/dec, until reaching a saturation point, at about $\approx 2000$ s.
\\
With that in mind, the final step examines the baseline sensitivity to $\omega_{ie}$ as a function of the averaging time. This allows us to gauge how waiting time affects precision, striking a balance between the desired accuracy and real-time deployment.
\\
Next, to gain a better understanding of the overall signal behavior across all three sensor channels, we express the signal magnitude in the i-th time step ($\Bar{\textbf{\textit{x}}}_i$) using the running version of the root mean square (RMS) formula:
\begin{align}
\mu (n)_{\, \text{RMS}} = \sqrt{ \frac{1}{n} \sum_{i=1}^{n} \big\| \, \Bar{\textbf{\textit{x}}}_i  \big\|^2 } \ .
\end{align}
\\
Fig.~\ref{fig:err_ma} illustrates the stabilization timeline of the baseline $\mu(t)$, along with its corresponding standard deviations ($\pm \sigma(t)$). The top subfigure presents the short-term error behavior, while the bottom one shows its asymptotic convergence over longer averaging periods, demonstrating how the inclusion of additional samples effectively cancels out the noise effects.
As shown, the mean estimate gradually converges to a value of 15.8 $^\circ$/hr after roughly 30 minutes, resulting in a systematic error of 0.8 $^\circ$/hr, equivalent to a 5.3\% relative error.
\\
In summary, this section illustrates the baseline capabilities of estimating the meaningful signal, which is interchangeably referred to as our baseline. While the sensor boasts low noise density, it still proves sensitive to emerging instabilities, gradually obscuring the faint rotation signal. 

\subsection{Hyperparameter selection}
Critical to the performance and generalization of deep learning models are the model weights. In the absence of a differentiable closed-form expression, a heuristic search becomes essential in the pursuit of optimality. Table~\ref{t:hyper} outlines various configurations determined heuristically, each successfully minimizing one or more of the following objectives: the number of parameters, inference latency, and estimation error.
\begin{table}[h]
\centering
\caption{Three suboptimal hyperparameters configurations.}
\renewcommand{\arraystretch}{1.8}
\begin{tabular}{c|c|c||c|c|c}
\shortstack{\# Layers $\times$\\hidden-size} & \shortstack{Batch\\-size} & \shortstack{Learn.\\-rate} & Params. & \shortstack{Latency\\time [ms]} & \shortstack{Estimation\\error [deg]} \\ \specialrule{1.25pt}{1pt}{1pt}
1 $\times$ 8 &  50 & 0.0005 & \textbf{9265} & \textbf{3.99} & 2.81$ \pm$ 0.71 \\ \hline
2 $\times$ 16 & 100 & 0.0015 & 42065 & 7.81 & 2.13 $\pm$ 0.30\\ \hline
2 $\times$ 24 & 100 & 0.0015 & 98965 & 26.62 & $\textbf{1.91}\pm\textbf{0.26}$ \\ \specialrule{1.25pt}{1pt}{1pt}
\end{tabular} \label{t:hyper}
\end{table}
\\
Given no computational limitation, the settings at the bottom of the table were conveniently chosen as our model weights. 

\subsection{Model performances} \label{sub_s:model}
\begin{figure}[h] 
\begin{center}
\includegraphics[width=0.5\textwidth]{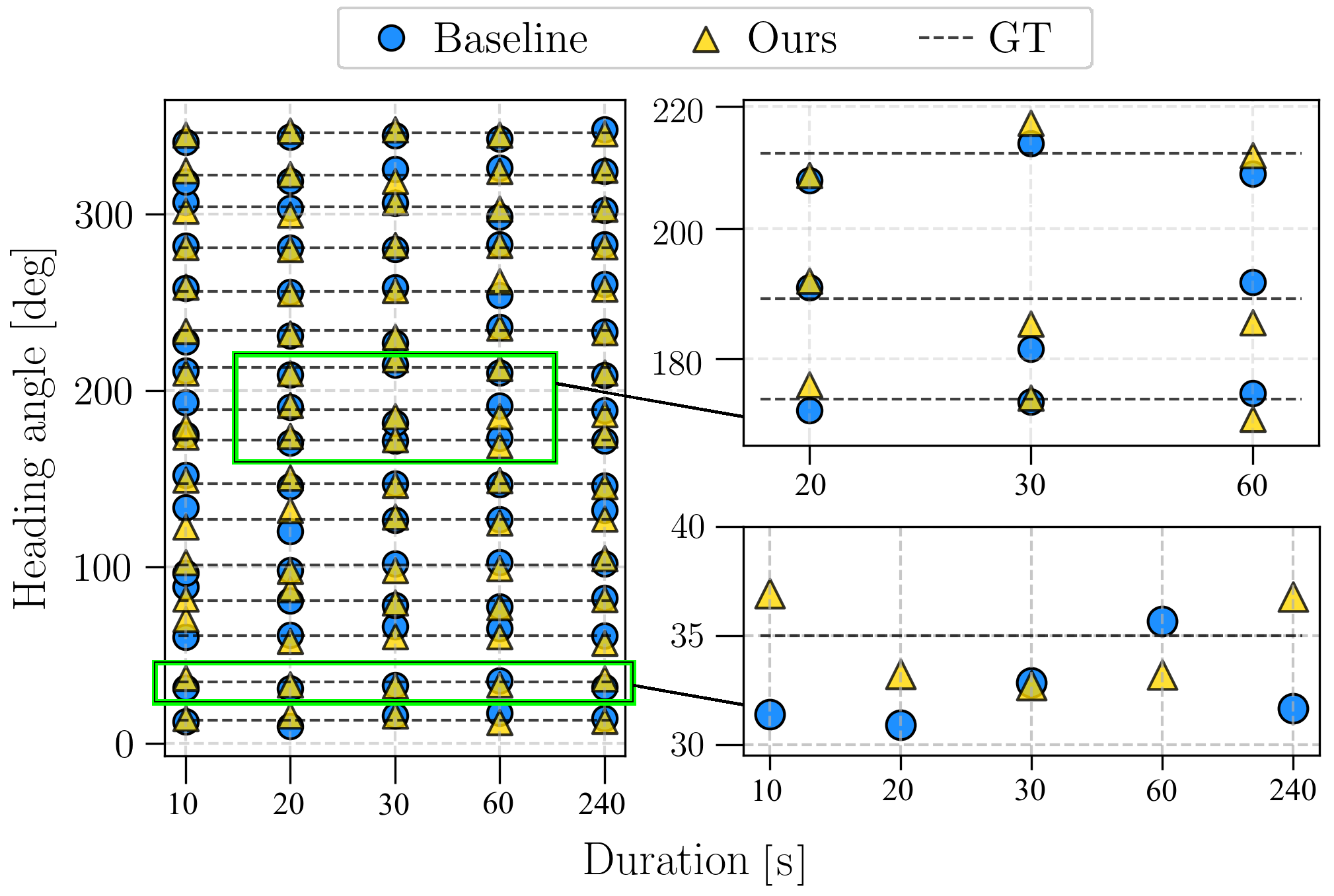}
\caption{Performance comparison over entire test batch.}
\label{fig:err_batch}
\end{center}
\end{figure}

\begin{figure}[t] 
\begin{center}
\includegraphics[width=0.5\textwidth]{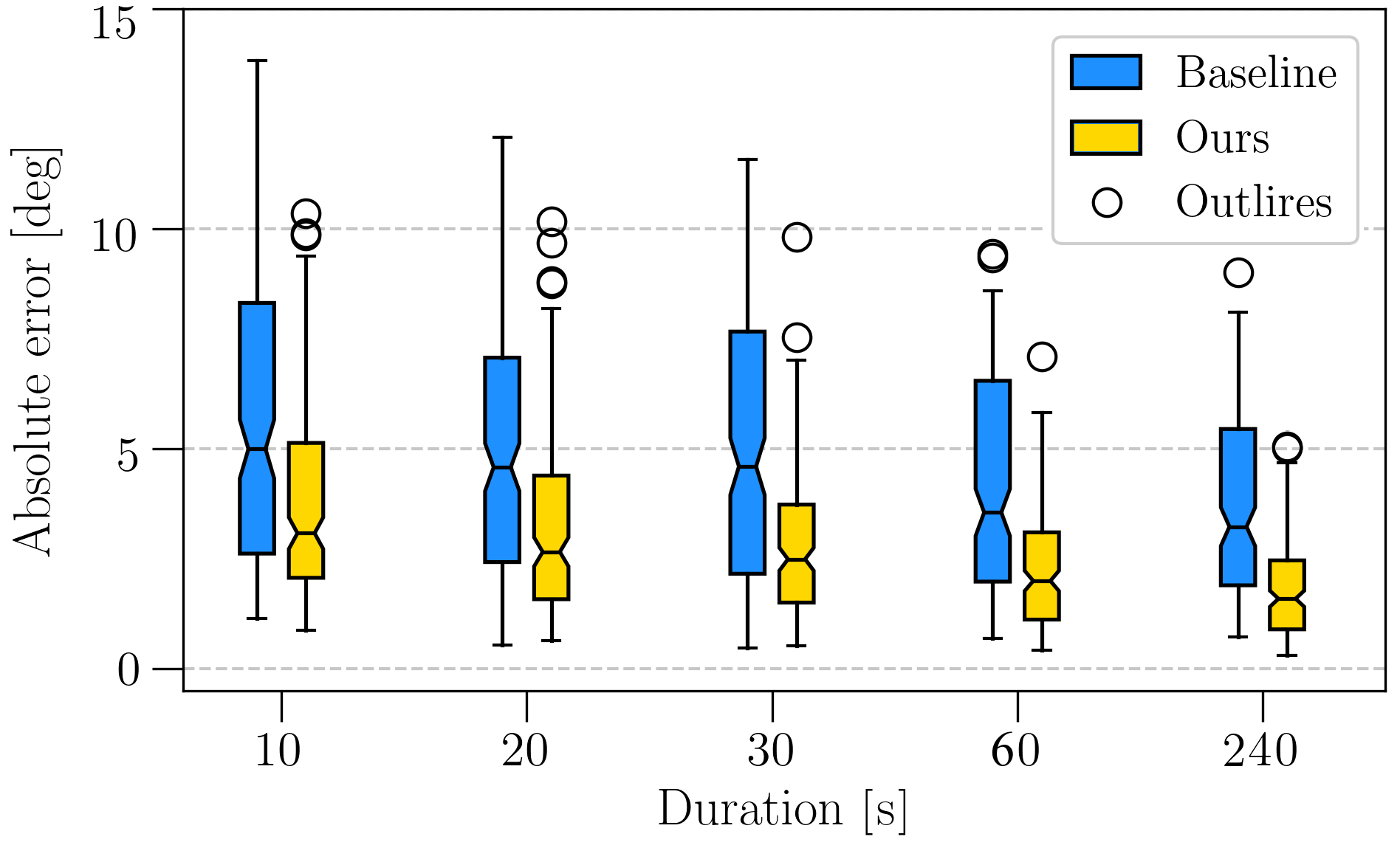}
\caption{Boxplot visualization of the models performances.}
\label{fig:err_dist}
\end{center}
\end{figure}

After learning about the inherent model limitations, we showcase our proposed approach and examine whether its gyrocompassing solution offers a competitive advantage\footnote{Data and source code are available @ \underline{\href{https://github.com/ANSFL/Learning-Based-MEMS-Gyrocompassing}{https://github.com/ansfl/LBGC}.}}.
\begin{figure}[h] 
\begin{center}
\includegraphics[width=0.5\textwidth]{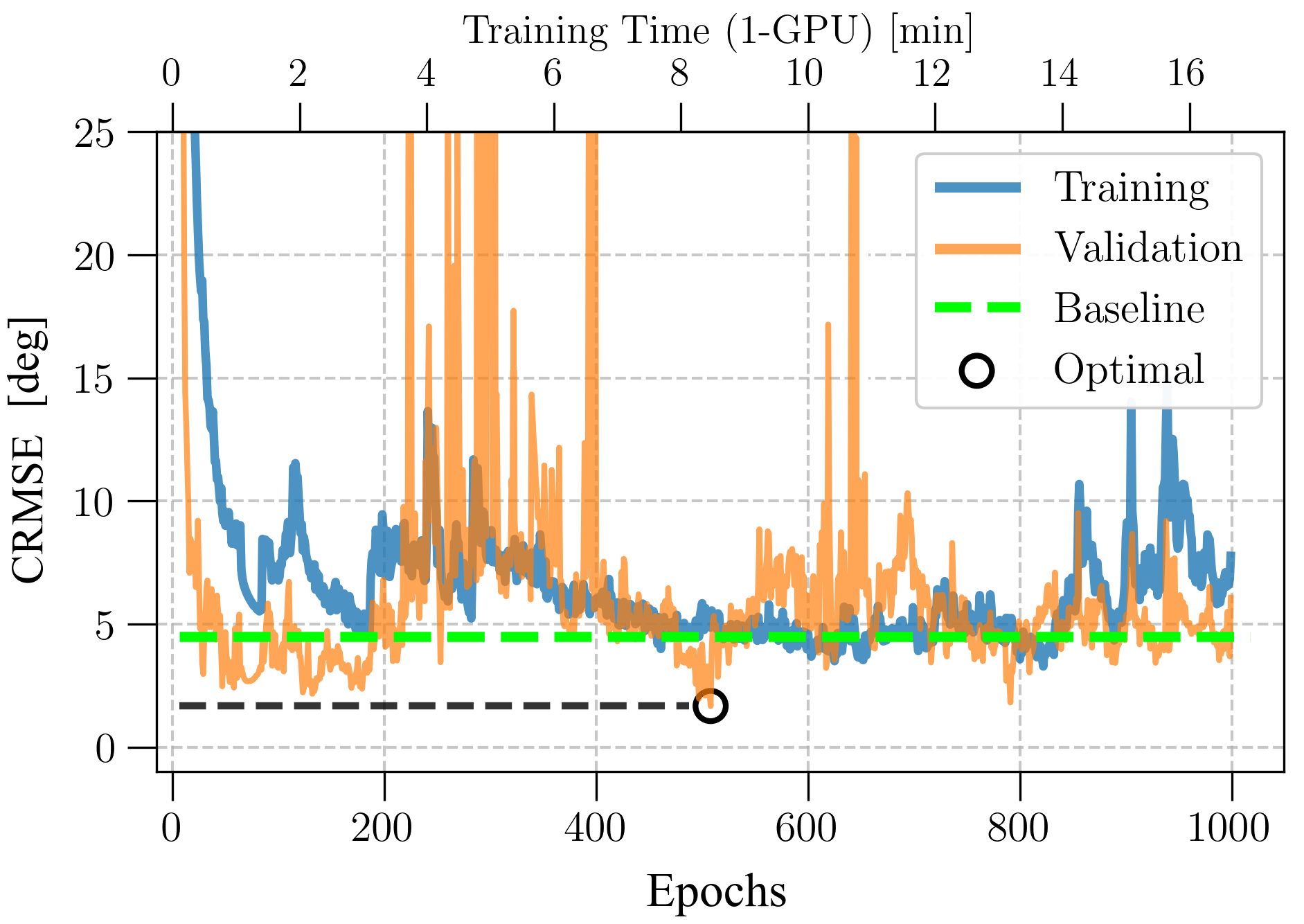}
\caption{The learning process of our proposed Bi-LSTM model.}
\label{fig:loss}
\end{center}
\end{figure} 
\\
Fig.~\ref{fig:loss} illustrates the learning progress as a function of the CRMSE, representing the square root of our loss ($\sqrt{\mathcal{L}_{\text{CMSE}}}$). Across 1000 training epochs, optimal performance is attained at the 510th epoch, reaching a minimum CRMSE of 1.88$^\circ$ per a 16-sample validation batch. However, it is also to suggest the minimum training epochs required to surpass the baseline.
The notable fluctuations in the validation curve can be attributed to the limited number of samples, which, unlike the training set, undergo no augmentation. 
\\
However, despite the remarkable improvement ratio of up to 50\%, better results are likely to be achieved across different executions-either due to more sensible hyperparameters, or data reshuffling. Following, the first evaluation task involves a 1-hour measurement, enabling continuous baseline estimation in the background, with emphasis on five different window sizes (10, 20, 30, 60, and 240 seconds) for a head-to-head comparison.
\begin{figure}[b] 
\begin{center}
\includegraphics[width=0.5\textwidth]{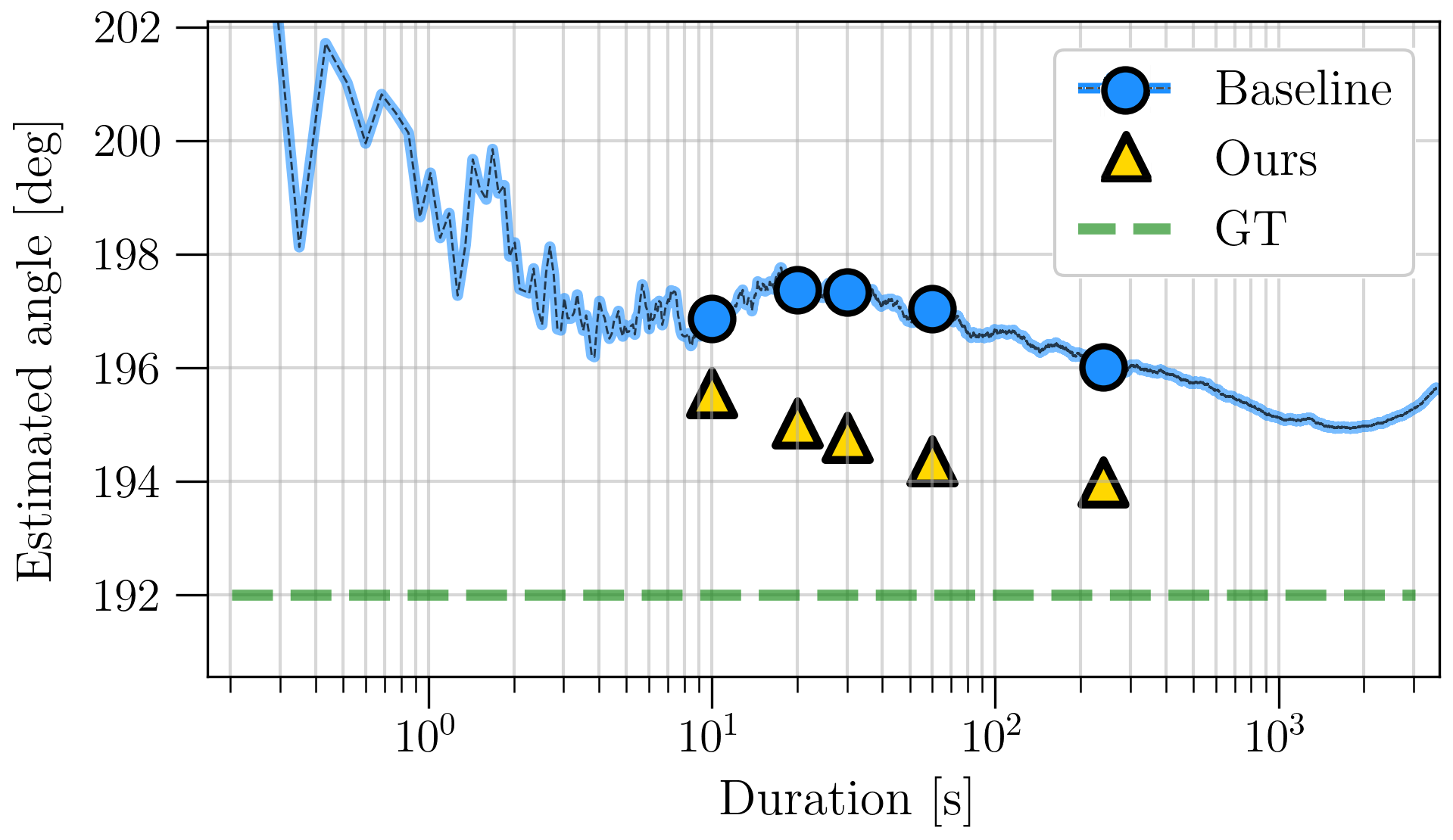}
\caption{Performance comparison over a single measurement.}
\label{fig:err_single}
\end{center}
\end{figure}
%
Fig.~\ref{fig:err_single} demonstrates this by comparing the baseline performances (blue bubbles) with our model estimates (yellow triangles), whereas the GT heading angle of $\psi_{\,\text{GT}}=192.1 ^\circ$, is provided for reference in green.
\\
As observed in many examples, but notably evident in this case, the noise presence induces bouncing estimates over short time scales ($\lesssim$ 20 s), thereby delaying the formation of a monotonic error trend. However, after around 2000 s, non-stationary error sources lead to what appears to be a saturation point, gradually degrading the baseline accuracy.
Traditionally, mitigating bias drift is achieved effectively through Kalman filtering, where it is formulated as an independent state variable. Nevertheless, this transition demands computational capacity, extra time reserves, and precise knowledge of initial conditions. 
\\
As a provisional conclusion, the observed ability to compensate for online drift may prove advantageous in unaided navigation scenarios (no filtering) or in reducing the required duration of fine alignment (with filtering). 
\\
A numerical context of that is given in Table~\ref{t:err_single}, comparing our model's performance with the baseline. The improvement ratios span from 31.7\% to 52.9\% across all five window sizes, highlighting an error limit that exceeds that of the baseline in both time and accuracy aspects. 
\begin{table}[h]
\centering
\caption{Error vs. window size from a single measurement.}
\renewcommand{\arraystretch}{1.8}
\begin{tabular}{c c|c|c|c|c|c|}
\multicolumn{2}{c}{} & \multicolumn{5}{c}{\textbf{Window size}} \\ \cline{3-7} 
 & & 10 [s] & 20 [s] & 30 [s] & 60 [s] & 240 [s] \\ 
\cline{3-7}\addlinespace\cline{2-7} 
\multirow{2}{*}{\rotatebox[origin=c]{90}{\textbf{Error}}} & \multicolumn{1}{|c|}{Baseline [deg]} & 4.77 & 5.26 & 5.23 & 4.94 & 4.01 \\ \cline{2-7}
    & \multicolumn{1}{|c|}{Ours [deg]} & 3.26 & 2.93 & 2.58 & 2.26 & 1.89 \\ 
    \cline{2-7} 
    & \multicolumn{1}{|c|}{Improvement [\%]} & \textbf{31.7} & \textbf{44.3} & \textbf{50.8} & \textbf{54.2} & \textbf{52.9} \\ \cline{2-7}
\end{tabular} \label{t:err_single}
\end{table} 
\\
Next, to gain a broader understanding, performance is assessed over an unseen test set comprising 16 samples. 
These are drawn from the 360$^\circ$ azimuthal circle at an average interval of about $22^\circ$, as illustrated in the three subfigures of Fig.~\ref{fig:err_batch}.
\\
The leftmost panel presents results across the entire test batch, offering a comprehensive view of the inference task across the same set of timestamps. The panels on the right zoom in on two specific regions, highlighting the relationship between effectiveness and duration. 
%
\\
To substantiate these assertions from a statistical standpoint, additional analysis spanning the temporal axis is presented in Fig.~\ref{fig:err_dist}. Given its limited size, the unaugmented test set may exhibit potential sensitivity to skewness and outliers. To address this concern, boxplot visualization is employed, offering an intuitive interpretation of both data locality and dispersion.
%
Each "box" defines the interquartile range (IQR), spanning from the first quartile (Q$_1$) to the third quartile (Q$_3$), while the median line (Q$_2$) is situated within this range. While individual outliers are marked with circles, the external lines ("whiskers"), extend to the farthest in-range values.
\\
A numerical summary of this is provided in Table~\ref{t:err_batch}, where two key parameters are extracted from the boxplot statistics, emphasizing their role in capturing time-dependent performance and facilitating a comparative perspective.
\begin{table}[h]
\centering
\caption{Error vs. window size from entire test set.}
\renewcommand{\arraystretch}{1.9}
\begin{tabular}{c c|c|c|c|c|c|}
\multicolumn{2}{c}{} & \multicolumn{5}{c}{\textbf{Window size}} \\ \cline{3-7} 
 & & 10 [s] & 20 [s] & 30 [s] & 60 [s] & 240 [s] \\ 
    \cline{3-7} 
    \addlinespace
    \cline{2-7} 
\multirow{2}{*}{\rotatebox[origin=c]{90}{\textbf{Median}}} & \multicolumn{1}{|c|}{Baseline [deg]} & 
5.06 & 4.73 & 4.68 & 3.67 & 3.41 \\ \cline{2-7}
    & \multicolumn{1}{|c|}{Ours [deg]} & 3.27 & 2.68 & 2.52 & 1.96 & 1.51 \\ \cline{2-7} 
    & \multicolumn{1}{|c|}{Improvement [\%]} & \textbf{35.2} & \textbf{43.3} & \textbf{46.2} & \textbf{46.6} & \textbf{55.7} \\ 
    \cline{2-7} \addlinespace \cline{2-7} 
\multirow{2}{*}{\rotatebox[origin=c]{90}{\textbf{IQR}}}  & \multicolumn{1}{|c|}{Baseline [deg]} & 5.62 & 4.63 & 5.66 & 4.27 & 3.72 \\ \cline{2-7}
    & \multicolumn{1}{|c|}{Ours [deg]} & 3.07 & 2.71 & 2.11 & 1.90 & 1.62 \\ \cline{2-7} 
    & \multicolumn{1}{|c|}{Improvement [\%]} & \textbf{45.4} & \textbf{37.8} & \textbf{62.7} & \textbf{55.5} & \textbf{56.5} \\ \cline{2-7}
\end{tabular} \label{t:err_batch}
\end{table} 
\\
As evident, the same trends observed previously are once again replicated across all 16 different samples, with improvement ratios ranging between 35.2\% and 55.7\%. 
%
Regardless of the model under consideration, both the median estimates and their corresponding IQRs exhibit a steady decrease, maintaining an approximate performance ratio of about one half. 
\\
One final insight that is worth dwelling on is the potential time reduction achieved by swapping between the error and time variables. Upon interpolation, the median errors from both models facilitate the regression of desirable performance thresholds, revealing the time-cost constraint for each error. 
\begin{table}[h]
\centering
\caption{Time-error trade-off across the entire test set.}
\renewcommand{\arraystretch}{1.9}
\begin{tabular}{c c|c|c|c|c|c|c|}
\multicolumn{2}{c}{} & \multicolumn{6}{c}{\textbf{Median Error}} \\ \cline{3-8} 
 & & 4.0$^\circ$ & 3.5$^\circ$ & 3.0$^\circ$ & 2.5$^\circ$ & 2.0$^\circ$ & 1.5$^\circ$ \\ 
    \cline{3-8} 
    \addlinespace
    \cline{2-8} 
\multirow{2}{*}{\rotatebox[origin=c]{90}{\textbf{Duration}}} & \multicolumn{1}{|c|}{Baseline [s]} & 50.6 & 78.3 & 203.1 & 984.2 & \textbf{N/A} & \textbf{N/A} \\ \cline{2-8}
    & \multicolumn{1}{|c|}{Ours [s]} & 5.7 & 7.22 & 13.8 & 31.1 & 47 & 248 \\ \cline{2-8} 
    & \multicolumn{1}{|c|}{Improvement [\%]} & \textbf{88.7} & \textbf{90.8} & \textbf{93.2} & \textbf{96.8} & - & - \\ \cline{2-8}
\end{tabular} \label{t:err_time}
\end{table} 
\\
As observed in Table~\ref{t:err_time}, while the cumulative errors set a lower bound on the baseline error (see Fig.~\ref{fig:err_single}), our model remains unaffected, consistently reducing the heading error. 
Adopting this interpretation emphasizes a remarkable time reduction by more than one order of magnitude ($\times$10), bringing significant capability and novelty to the challenging gyrocompassing task. 
These aspects hold utmost importance, suggesting that new error limits can now be achieved through learning, whereas conventionally, they remain \underline{unattainable}, even with an indefinite sampling period. 

\section{Discussion} \label{sec:disc}
After a comprehensive analysis of the experimental results, our work reveals three noteworthy insights:
\begin{enumerate}
    \item \textbf{Accuracy and precision}: Upon scrutinizing the error distributions generated by the models' predictions, the well-trained model exhibited lower systematic errors, coupled with a diminished level of dispersion. This observation implies a heightened level of confidence in its estimates.
    \item \textbf{Asymptotic consistency}: With longer measurement durations, the estimation error decreases. The estimates consistently converged towards the true parameter being estimated, a principle applicable to both models.
    \item \textbf{Adaptability}: The learning-based approach demonstrates an impressive capacity to adapt to complex relationships within the sensor, effectively compensating for intricate errors. This capability proves particularly advantageous in real-world scenarios, as it also recognizes the analytical limitations inherent in predefined model assumptions.
\end{enumerate}

\subsection{Limitations of this study}
Considering the observed strengths emerging from this study, certain limitations should be acknowledged as well: 
\begin{enumerate}
    \item \textbf{Sensitivity threshold}: The proposed method's validity was tested on a sensor with a noise floor in the vicinity of Earth's rotation amplitude. Despite not being originally designed for gyrocompassing, the sensor's relatively low SNR was successfully compensated during the training phase, resulting in a 3$^\circ$ heading error. 
    \\
    However, it is essential to note that low-cost MEMS IMUs, such as those found in our mobile phones, characterized by an extremely low sensitivity to $\omega_{ie}$, are not likely to benefit from this approach.
    \item \textbf{Stationary conditions}: Conducting the experiment under entirely stationary conditions aims to emphasize error patterns, identify them, and, most importantly, allows the model to capture their time dependencies In realistic scenarios, environmental disturbances and platform dynamics are expected. These factors are likely to overpower the weak error patterns, thus requiring an appropriate adjustment before the direct application of the proposed approach.
\end{enumerate}

\subsection{Future study}
Considering the promising outcomes of the study and acknowledging its limitations, we suggest two main avenues for future exploration:
\begin{enumerate}
    \item \textbf{Sensors diversity}: Explore the potential improvement rates over sensors of different qualities, empirically examining which category benefits most from our approach, ranging from low-cost to high-cost IMUs.
    \item \textbf{Dynamic environments}: Extend the proposed solution into aerial or marine scenarios, where complex dynamics may occur, and assess the generalization limits of the model.
\end{enumerate}

\section{Conclusion} \label{sec:conc}
This study aimed at investigating the viability of mid-tier IMUs for gyrocompassing, a task that poses significant challenges due to the borderline sensitivity of these sensors. Following a comprehensive review of the prominent literature, the relevant background was provided, from which the proposed approach was presented, justified, and an appropriate experiment was designed.
The results demonstrated the significant capability of the proposed Bi-LSTM model in achieving better predictions while capturing intricate error relationships and compensating for them during testing. This indicates the inherent potential of this novel approach, achieved solely through software modification, resulting in significantly improved initial conditions. 
From a practical standpoint, this also implies that the gyrocompassing capability, once reserved for high-end instruments only, is slowly becoming accessible in off-the-shelf products, making it available to the average consumer. These findings hold great practical significance, particularly in robotics, tactical and autonomous platforms, where the capability for rapid deployment or alignment is considered a game-changer.

\bibliographystyle{ieeetr}
\bibliography{Ref}
%
\end{document}